\documentclass{article}

\usepackage{microtype}
\usepackage{graphicx}
\usepackage{subfigure}
\usepackage{booktabs} % for professional tables

\usepackage{hyperref}

\usepackage[accepted]{icml2019}

\icmltitlerunning{Deep Generative Quantile-Copula Models for Probabilistic Forecasting}

\usepackage{graphicx}
\usepackage{amssymb,amsmath}
\usepackage{longtable}
\usepackage{xtab}
\makeatletter
\def\longtable{\@ifnextchar[\newlongtable@i \newlongtable@ii}
\def\newlongtable@i[#1]{%
\renewcommand{\endhead}{\ignorespaces}
\begin{table*}[t]
\centering
\caption{Experiment Metrics. See texts for explanation.}
\xtabular[#1]}
\def\newlongtable@ii{%
\renewcommand{\endhead}{\ignorespaces}
\begin{table*}[t]
\centering
\caption{Experiment Metrics. See texts for explanation.}
\xtabular}
\def\endlongtable{\endxtabular
\end{table*}}
\makeatother

\begin{document}

\twocolumn[
\icmltitle{Deep Generative Quantile-Copula Models for Probabilistic Forecasting}

\begin{icmlauthorlist}
\icmlauthor{Ruofeng Wen}{fcst}
\icmlauthor{Kari Torkkola}{fcst}
\end{icmlauthorlist}

\icmlaffiliation{fcst}{Forecasting Data Science, Amazon}

\icmlcorrespondingauthor{Ruofeng Wen}{ruofeng@amazon.com}
\vskip 0.3in
]

\printAffiliationsAndNotice{} 

\begin{abstract}
We introduce a new category of multivariate conditional generative
models and demonstrate its performance and versatility in probabilistic
time series forecasting and simulation. Specifically, the output of
quantile regression networks is expanded from a set of fixed quantiles
to the whole Quantile Function by a univariate mapping from a latent
uniform distribution to the target distribution. Then the multivariate
case is solved by learning such quantile functions for each dimension's
marginal distribution, followed by estimating a conditional Copula to
associate these latent uniform random variables. The quantile functions
and copula, together defining the joint predictive distribution, can be
parameterized by a single implicit generative Deep Neural Network.
\end{abstract}

\hypertarget{introduction}{%
\section{Introduction}\label{introduction}}

We start by describing implicit generative models of which the proposed
model is an instance, then motivations in the frontier of the challenges
in probabilistic time series forecasting and simulation.

\textbf{Implicit Generative Models} Consider learning a conditional
joint\footnote{Throughout this text, \emph{conditional} means
  conditioning on \(\mathbf{x}\), the features, not on part of
  \(\mathbf{y}\) itself. In contrast, \emph{marginal} and \emph{joint}
  are used regarding to \(\mathbf{y}\) itself only. All the formulation
  holds trivially for the unconditional case without \(\mathbf{x}\). For
  notation simplicity, \emph{conditional} may be omitted when there is
  no ambiguity.} distribution \(p(\mathbf{y}|\mathbf{x})\) from data
observations, where \(\mathbf{y}\) is a \(d\)-dimensional target random
vector and \(\mathbf{x}\) is a feature vector, through an implicit
generative model (IGM) \(\mathbf{y} = g(\mathbf{x},\mathbf{z})\). Here
\(\mathbf{z}\) is some latent random noise used to capture all the
underlying randomness in \(\mathbf{y}\) given \(\mathbf{x}\), through a
deterministic generator function \(g(\cdot)\), parameterized by a deep
neural net. The unsupervised version, without covariates \(\mathbf{x}\),
is the more popular formulation in literature with state-of-the-art
solutions in image and text generation. Generative adversarial networks
(GAN, \href{https://arxiv.org/abs/1406.2661}{Goodfellow et al, 2014})
map \(\hat{\mathbf{y}} = g(\mathbf{z})\) by training \(g(\cdot)\) to
fool a classifier telling real \(\mathbf{y}\) from the generated
\(\hat{\mathbf{y}}\). GANs suffer from stability issues in training and
complex constraints of the critic function in its variants (e.g.~WGAN,
\href{https://arxiv.org/abs/1701.07875}{Arjovsky et al, 2017}). The
learning principle of GAN-style IGMs is \emph{comparison by samples}
(\href{https://arxiv.org/abs/1610.03483}{Mohamed and Lakshminarayanan,
2016}), basically comparing the empirical distribution between an
observations set and the generated samples set,
e.g.~adversarial/f-divergence/moment-matching losses. This approach
falls short in conditional generative modeling: the specific context
\(\mathbf{x}\) of an observation usually appears only once in the
dataset, thus it is difficult to do sub-population comparison. This
issue naturally leads to the use of \emph{proper scoring rules}
(\href{https://www.stat.washington.edu/raftery/Research/PDF/Gneiting2007jasa.pdf}{Gneiting
and Raftery, 2007}) which compare a single observation against a
predictive distribution. Flow-based generative models
(\href{https://arxiv.org/abs/1410.8516}{Dinh et al,
2014}/\href{https://arxiv.org/abs/1605.08803}{2016}) learn by maximizing
log-likelihood, the most common scoring rule. Flow-based models restrict
the mapping \(\mathbf{y} = g(\mathbf{z})\) to be invertible and the
Jacobian determinant
\(|\mathrm{d}g^{-1}(\mathbf{y})/\mathrm{d}\mathbf{y}|\) to be easy to
compute. This significantly simplifies the likelihood inference
\(p(\mathbf{y}) = p(\mathbf{z})|\mathrm{d}\mathbf{z}/\mathrm{d}\mathbf{y}|\)
with \(\mathbf{z} = g^{-1}(\mathbf{y})\). Such restricted network layers
are however less expressive, especially when \(\mathbf{x}\) is present.
Autoregressive models (\href{https://arxiv.org/abs/1502.03509}{Germain
et al, 2015}, \href{https://arxiv.org/abs/1609.03499}{Van Den Oord et
al, 2016}) remove the explicit need of \(\mathbf{z}\) by
self-decomposing
\(p(\mathbf{y}) = p(y_1,\cdots,y_d) = p(y_1)\prod_{i}p(y_{i+1}|y_i,\cdots,y_1)\)
and generate \(\hat{\mathbf{y}}\) by recursively drawing and feeding
one-step-ahead samples, resulting in an expressive
univariate-to-multivariate likelihood parameterization but also creating
issues in order picking, error accumulation and heavy sampling
computation. Another major competitor of IGMs is latent variable model,
particularly variational auto-encoder (VAE,
\href{https://arxiv.org/abs/1312.6114}{Kingma and Welling, 2014}), where
\(p(\mathbf{y})\) is characterized by
\(\int{p(\mathbf{y}|\mathbf{z})p(\mathbf{z})}\mathrm{d}\mathbf{z}\).
Such models suffer from intractable integrals and limitations with
prescribed families of distributions.

\textbf{Probabilistic Forecasting} The probabilistic time series
forecasting problem can be formulated as learning
\(p(y_{t+d},\cdots,y_{t+1}|y_{:t},\mathbf{x})\) where \(y_{:t}\) denotes
the observed series before time \(t\). Multi-horizon quantile forecaster
(MQ-RNN/CNN, \href{https://arxiv.org/abs/1711.11053}{Wen et al, 2017})
combines multi-horizon forecasts, quantile regression and
sequence-to-sequence architecture, and predicts multiple quantiles for
each future horizon. \href{https://arxiv.org/abs/1711.11053}{Wen et al,
2017} demonstrated that MQ-forecasters have superior accuracy over
autoregressive and parametric likelihood deep forecasting models,
e.g.~variants of DeepAR
(\href{https://arxiv.org/abs/1704.04110}{Flunkert et al, 2017}), and
also over classical forecasting methods in a previous public
competition.
\href{https://milets18.github.io/papers/milets18_paper_14.pdf}{Madeka et
al, 2018} further showed that common deep generative models, including
VAE, GAN, Bayesian Dropout (\href{https://arxiv.org/abs/1506.02142}{Gal
and Ghahramani, 2015}) and WaveNet
(\href{https://arxiv.org/abs/1609.03499}{Van Den Oord et al, 2016}) all
have large gaps towards the accuracy of MQ-forecasters. Despite the
success, there are several missing pieces in the framework: (1) Plain
MQ-forecaster outputs the marginal distributions of each future horizon
\(p(y_{t+i}|y_{:t},\mathbf{x})\), \(i = 1,\cdots,d\), not the joint
distribution \(p(y_{t:}|y_{:t},\mathbf{x})\), due to the univariate
nature of quantiles. One workaround is the Mesh Approach, at the cost of
potential statistical inconsistencies in the forecast distribution
(detailed in
\protect\hyperlink{mesh-gamma-and-evaluation-metrics}{Section 4.1}). (2)
MQ-forecaster predicts a pre-defined set of quantiles only. This is
computationally inefficient if the set is large, and leads to complex
ad-hoc procedures in interpolation or parametric fitting (detailed in
\protect\hyperlink{mesh-gamma-and-evaluation-metrics}{Section 4.1}). (3)
It is not a generative model, and thus cannot meet certain application
requirements (e.g.~demand simulation for inventory control and
reinforcement learning). (4) It does not have a native way to express
cross-series association (e.g.~correlation among products/business-group
series), so each time series has to be treated independently. With such
limitations, MQ-forecaster is not a \emph{complete} solution to
probabilistic forecasting and simulation.

\textbf{Our Contribution} In this paper, we propose a new kind of deep
implicit generative model. It works separately as a general approach
with advantages in conditional modeling over existing choices. Plugging
it into the MQ-forecaster can fill in all the above missing pieces,
yielding a fully generative joint forecast distribution for time series
simulation and anomaly detection, while maintaining accuracy.
Specifically, we design a conditional generative \emph{Quantile-Copula}
framework, parameterized by a single deep neural network. Unlike other
implicit generative models, where a set of random noises is directly
translated into target distribution through black-box transformations,
we focus on decoupling the complex marginal shapes (quantile function)
and the pure joint association (copula). In terms of optimization, such
decoupling enables the use of Quantile Loss, a computationally simple
piecewise linear loss function, also a proper scoring rule, that can
reliably learn arbitrarily complex nonparametric conditional
distributions. In terms of statistical modeling, this work is also a
practical attempt to formulate a Multivariate Quantile Regression. The
proposed deep Quantile-Copula model suits applications that require
accurate and calibrated characterization of each target random variable
in multi-target learning, also with the need to simulate or infer the
joint distribution of target vector. For example, image and text data
would benefit less but time series and network data will gain more since
the value at each time point or graph node matters.

In \protect\hyperlink{background-and-building-blocks}{Section 2} we
introduce some methods as building block, then describe the
Quantile-Copula model, as well as its usage in time series forecasting
in \protect\hyperlink{generative-quantile-copula-model}{Section 3}.
Experiments with Amazon Demand Forecast problem is presented in
\protect\hyperlink{experiment-amazon-demand-forecasting}{Section 4}.
Related work and future work are in
\protect\hyperlink{discussion}{Section 5}

\hypertarget{background-and-building-blocks}{%
\section{Background and Building
Blocks}\label{background-and-building-blocks}}

\hypertarget{quantile-regression-as-a-generative-model}{%
\subsection{Quantile Regression as a Generative
Model}\label{quantile-regression-as-a-generative-model}}

A classical Quantile Regression
(\href{https://pdfs.semanticscholar.org/a3cd/bfbba2ef3ce285980edc1213a4ac56f05bb1.pdf}{Koenker
and Gilbert, 1978}) predicts the conditional \(u\)th quantile
\(y^{(u)}\) given covariates \(\mathbf{x}\) and a fixed \emph{quantile
index} \(u\in [0,1]\), such that \(P(y \leq y^{(u)}|\mathbf{x}) = u\).
The model is trained by minimizing the total Quantile Loss (QL; also
known as pinball or check loss):
\[QL_u(y,\hat{y}^{(u)}) = u(y-\hat{y}^{(u)})_{+} + (1-u)(\hat{y}^{(u)}-y)_{+}\]
where \((\cdot)_{+}=\max(0,\cdot)\). The model can be parameterized by
any function: \(y^{(u)} = g_u(\mathbf{x})\). In the classical case of
linear function, one \(g_u(\cdot)\) is fitted for each given \(u\)
needed by the application. However, using an expressive deep neural net
as a complex non-linear function approximator,
\href{https://arxiv.org/abs/1806.06923}{Dabney et al, 2018} suggested an
efficient setup: \(y^{(u)} = g(u,\mathbf{x})\), i.e.~the quantile index
\(u\) is an input to the neural net as a feature, and also as the weight
in the loss function when training. \(u\) effectively tells the neural
net which quantile to generate when predicting. See Figure 1 (a) and
(b).

Such model is essentially learning the conditional Quantile Function
\(y = Q(u|\mathbf{x})\), which is the inverse of the conditional
distribution function \(F(y|\mathbf{x})\), i.e. \(Q = F^{-1}\), if
\(F(\cdot)\) is strictly monotonic. While any distribution function maps
a random variable following it to a uniform random variable, the
quantile function does the opposite: it maps a latent uniform random
variable (with the interpretation of being a quantile index, when
instantiated) to the target random variable. Thus the learned
\(g(u,\mathbf{x})\) is a univariate random number generator for \(y\)
given \(\mathbf{x}\), if \(u \sim U(0,1)\). In fact, during training,
\(u\) can be drawn from \(U(0,1)\) independently in each epoch, to pair
with each observation of \((\mathbf{x},y)\). In this way, the model
still converges to minimizing the expected QL across \(u \sim U(0,1)\)
(or the \emph{Quantile Divergence} as named by
\href{https://arxiv.org/abs/1806.05575}{Ostrovski et at, 2018}), given
there are enough epochs. This is far more efficient than computing QL at
all possible values of \(u\) for every sample.

\hypertarget{marginal-multi-quantile-model}{%
\subsection{Marginal Multi-Quantile
Model}\label{marginal-multi-quantile-model}}

If the target \(\mathbf{y}\) is \(d\)-dimensional,
\href{https://arxiv.org/abs/1711.11053}{Wen et al, 2017} and
\href{https://www.sciencedirect.com/science/article/pii/S0957417417300726}{Xu
et al, 2017} showed that all of the marginal quantiles can be
efficiently predicted by a neural net with matrix output
\(\mathbf{Y}_{d\times m}^{(\mathbf{u})} = [y_i^{(u_j)}]_{i,j}\), given
the fixed list of \(m\) quantile indices. Adopting the same generative
aspect as described above, this multi-quantile model can be seen as a
marginally generative model:
\[\mathbf{y}^{(\mathbf{u})} = (y_{1}^{(u_1)},\cdots,y_{d}^{(u_d)}) = g(\mathbf{u},\mathbf{x})\]

where \(\mathbf{u} \in [0,1]^d\) is a vector of quantile indices for
each element of \(\mathbf{y}\). The exact form of \(g(\cdot)\) can vary.
For example, in order of descending complexity: it could be \(d\)
different functions with the same covariates
\(y_i^{(u_i)} = g_i(u_i,\mathbf{x})\), or one function with different
parameters \(g(u_i,\mathbf{x};\theta_i))\), or one function with shared
parameters but split feature embeddings \(g(u_i,c_i(\mathbf{x}))\),
where \(c_i(\mathbf{x})\) is the \(i\)th target-related
contexts/conditioners extracted from all features. We adopt the last
parameterization in this text, while all discussion holds for others.
See Figure 1 (c). Similar to the univariate case, by setting \(u_i\) to
a specific value, the corresponding quantile prediction is obtained. And
by drawing \(u_i \sim U(0,1)\) we can generate the \emph{marginal}
target distribution. Note that there is no association among
\(u_1,\cdots,u_d\).

\hypertarget{copula-and-gaussian-copula}{%
\subsection{Copula and Gaussian
Copula}\label{copula-and-gaussian-copula}}

The joint cumulative distribution function of a set of marginally
\(U(0,1)\) random variables is called a \emph{Copula}, denoted by
\(C(\mathbf{u})\). By Sklar's Theorem
(\href{https://pdfs.semanticscholar.org/feed/9337c89730250b11ab722742cf27e94c335a.pdf}{Sklar,
1959}), every multivariate distribution function can be decomposed into
its marginals and a unique copula:
\[F(\mathbf{y}) = \prod_{i}u_i \cdot C(u_1,\cdots,u_d)\]

where \(u_i = F_i(y_i)\) and \(y_i \sim F_i(y)\). A copula characterizes
the association within the latent random vector in the normalized space,
decoupled from the possibly complex marginal-specific distributions.

One expressive family of copula is the Gaussian Copula. Let the standard
normal CDF be \(\Phi(\cdot)\), then a Gaussian copula for a random
vector \(\mathbf{u}\) is a distribution parameterized by a
\(d\)-by-\(d\) correlation matrix \(\mathbf{R}\), such that
\(\Phi^{-1}(\mathbf{u}) \sim N(\mathbf{0},\mathbf{R})\). That is, a
Gaussian copula assumes that the random vector \(\mathbf{u}\) is the
probability integral transform of a multivariate normal distribution
with zero mean and a correlation matrix. Given \(\mathbf{R}\),
generating samples \(\mathbf{u}\) from noise \(\mathbf{z}\) is simple.
One can draw \(d\) independent standard normal samples
\(\mathbf{z} = (z_1,\cdots,z_d)^\intercal \sim N(\mathbf{0},\mathbf{I})\),
multiply by the Cholesky lower-triangle matrix \(\mathbf{L}\) (s.t.
\(\mathbf{L}\mathbf{L}^\intercal = \mathbf{R}\)) to add association:
\(\mathbf{z}^{*} = \mathbf{L}\mathbf{z}\), and finally
\(\mathbf{u} = \Phi(\mathbf{z^*})\). See Figure 1 (d). During the
sampling (and learning, detailed later) \(\mathbf{R}\) is not explicitly
needed, and \(\mathbf{L}\) can be used to parameterize the same copula,
without the need of computing Cholesky decomposition. The simplicity in
drawing samples and conditioning contexts through \(\mathbf{L}\) in
neural nets is the main reason we found Gaussian copula favorable over
alternatives like empirical copula. However, the Gaussian constraint on
copula is not a necessity. See \protect\hyperlink{discussion}{Section 5}
for discussion.

\begin{figure}
\centering
\includegraphics[width=2.79167in,height=3.48958in]{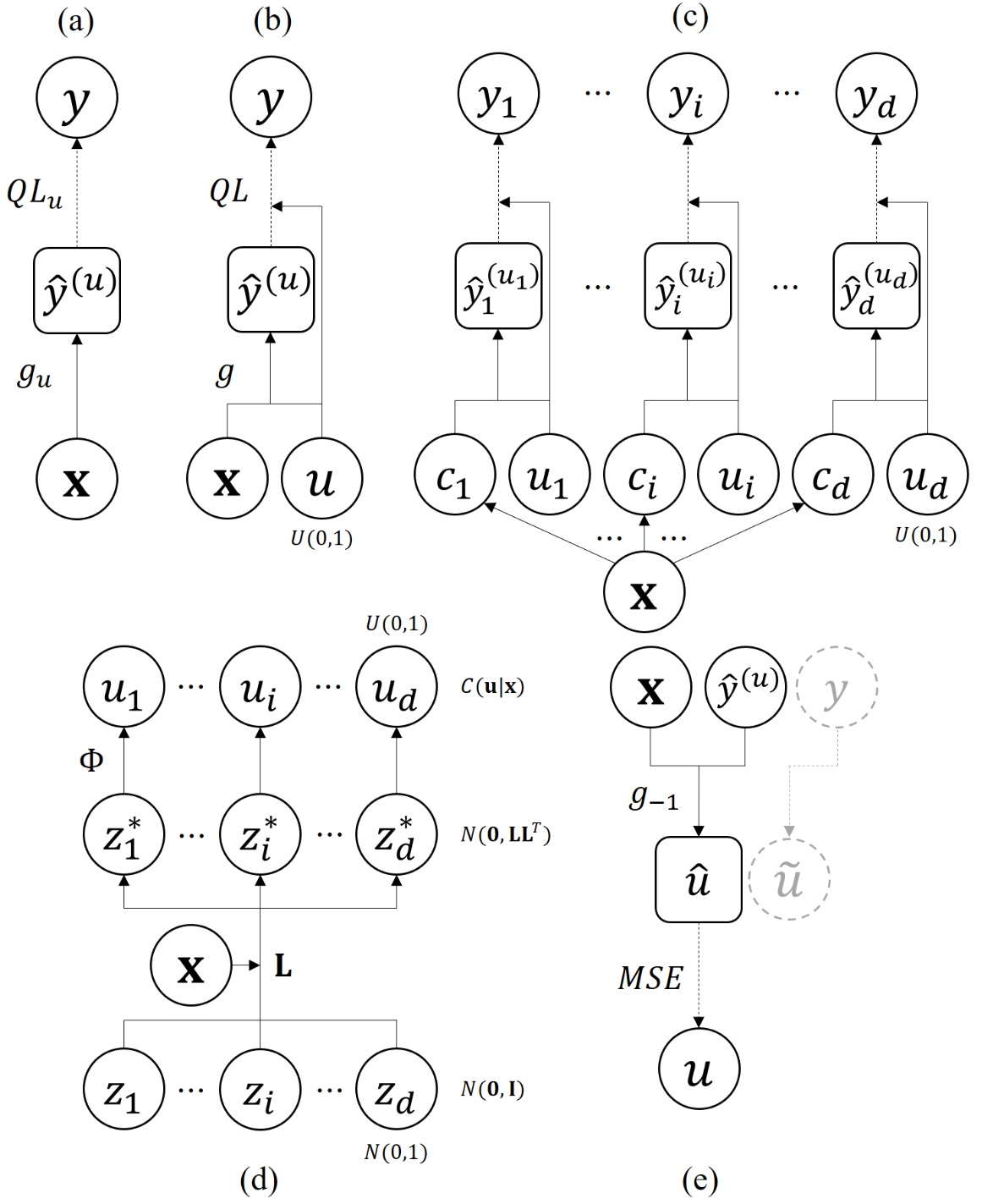}
\caption{Computational graphs and variable notations used in this paper.
Solid arrow indicates forward computation with possibly multiple layers,
and dashed arrow is the loss function linking prediction and truth. (a)
Quantile Regression; (b) Generative Quantile Model; (c) Generative
Multi-Quantile Model with a specific parameterization; (d) Generative
conditional Gaussian Copula; (e) Inverse MLP, shown for one target. Grey
nodes are the information flow during copula inference. Stacking (c)
over (d) leads to the proposed Quantile-Copula model. Notations: we use
\(\mathbf{y}\) for true targets, \(\hat{\mathbf{y}}\) for generated
predictions/samples, \(\mathbf{x}\)/\(\mathbf{c}\) for
features/contexts, \(\mathbf{u}\) for quantile indices of
\(\hat{\mathbf{y}}\) given \(\mathbf{x}\), \(\mathbf{z}\) the
independent random noises, \(\mathbf{z^*}\) the associated noises,
\(\hat{\mathbf{u}}\)/\(\tilde{\mathbf{u}}\) the estimated quantile
indices from \(\hat{\mathbf{y}}\)/\(\mathbf{y}\) given \(\mathbf{x}\).
See text for detailed discussion.}
\end{figure}

\hypertarget{generative-quantile-copula-model}{%
\section{Generative Quantile-Copula
Model}\label{generative-quantile-copula-model}}

We showed in the previous section that both quantile regression and
copula modeling can be rephrased as generative models. It is
straight-forward to combine them (by stacking Figure 1 (c) over (d)): a
copula can convert a set of independent random noises into a correlated
and marginally uniform random vector, then a series of quantile
functions can be element-wise applied to this vector, interpreted as
quantile indices, resulting in a sample conditioned on contexts.
Formally, for a random vector pair \((\mathbf{x},\mathbf{y})\), the
general Quantile-Copula model is:
\[\mathbf{y} = g_Q(\mathbf{u},\mathbf{x})\]
\[\mathbf{u} = g_C(\mathbf{z},\theta(\mathbf{x}))\]

where \(g_Q(\cdot)\) is the multi-target quantile function,
\(\mathbf{u}\) is the marginal quantile index vector for the
corresponding target vector \(\mathbf{y}\) given \(\mathbf{x}\),
\(g_C(\cdot)\) is the copula generator function, parameterized by
\(\theta\), to convert independent random noises \(\mathbf{z}\) to the
desired sample from the copula \(C_{\theta}(\mathbf{u}|\mathbf{x})\).
Essentially, this is a generator from noise \(\mathbf{z}\) to target
\(\mathbf{y}\) given contexts \(\mathbf{x}\), with a layer of
intermediate latent variables \(\mathbf{u}\) as the quantile indices of
each target. In this text, the following parameterization for \(g_Q\)
and \(g_C\) is used: \[ y_i = g(u_i,c_i(\mathbf{x})) \; \forall i \]
\[\mathbf{u} = \Phi(\mathbf{L}(\mathbf{x})\mathbf{z}) \]

where \(g(\cdot)\) acts as a universal conditional quantile function for
all targets\footnote{The choice of this parameterization is solely due
  to the fact that we are using time series as the main application, so
  all targets are intrinsically the same variable under different
  temporal contexts. Alternatives can be used if targets are
  heterogeneous and the difference cannot be fully characterized by
  contexts.}, and \(c_i(\cdot)\) returns target-specific feature
embedding as contexts. \(\mathbf{L}(\cdot)\) outputs a \(d\)-by-\(d\)
lower triangle matrix with a positive diagonal and unit row-norms (so
that \(\mathbf{L}\mathbf{L}^\intercal\) is a correlation matrix) based
on \(\mathbf{x}\), and \(\mathbf{z} \sim N(\mathbf{0},\mathbf{I})\). All
functions above are parameterized by either Multi-Layer Perceptrons
(MLPs) or structured deep nets (e.g.~for time series, \(c_i(\cdot)\) and
\(\mathbf{L}(\cdot)\) could be recurrent or convolutional nets over
sequential features).

\hypertarget{learning}{%
\subsection{Learning}\label{learning}}

The quantile part and the copula part of the model, including their loss
functions, are decoupled by the intermediate \(\mathbf{u}\). This
indicates that we can actually learn the quantile functions first, by
drawing \(\mathbf{u}\) from independent \(U(0,1)\) to pair with each
observation. Then the copula can be learned to associate \(\mathbf{u}\).
This two-phase training is favorable because many applications need the
quantile part only and the learning of the more difficult copula part
can be stabilized with well initialized quantile functions. For the
quantile part, the expected quantile loss needs to be minimized:
\[ l_1 = \mathbb{E}_{(\mathbf{x},\mathbf{y})}\mathbb{E}_{\mathbf{u}}[QL_{\mathbf{u}}(\mathbf{y},\hat{\mathbf{y}})] = \mathbb{E}_{(\mathbf{x},\mathbf{y},\mathbf{u})}[QL_{\mathbf{u}}(\mathbf{y},\hat{\mathbf{y}})] \]

where \(\hat{\mathbf{y}}\) is the output of \(g_Q(\cdot)\) and we abuse
the notation of \(QL(\cdot)\) as an apply-element-wise-then-sum
function. For learning the Gaussian Copula, maximum likelihood is used.
A prerequisite of computing the likelihood is to infer the latent
variable \(\mathbf{u}\) given the truth \(\mathbf{y}\). Such reverse
mapping can be done efficiently within the neural network. There are two
common solutions. One is to enforce invertibility: the choice of neural
structure in \(g(\cdot)\) is restricted to Flows, a set of invertible
layers; the other is to mimic the concept of auto-encoder: if
\(y_i = g(u_i,c_i(\mathbf{x}))\) is an MLP, then a structurally similar
\emph{inverse} MLP \(u_i = g_{-1}(y_i,c_i(\mathbf{x}))\) can learn this
inversion. In this text we pursue the latter because Flows significantly
restrict how contexts can be put into the net and thus limit the
expressiveness of \(g(\cdot)\). Also the inverse MLP works seamlessly in
our setting: it can be trained simultaneously with the forward MLP, with
the data pairs \((u_i \stackrel{g}{\to} \hat{y}_i)\) as ground truths,
available for free during the training. See Figure 1 (e). Let
\(\hat{u}_i = g_{-1}(\hat{y}_i,c_i(\mathbf{x}))\), the inverse
reconstruction loss is:
\[ l_2 = \mathbb{E}_{(\mathbf{x},\mathbf{y},\mathbf{u})}||\hat{\mathbf{u}} - \mathbf{u}||^2 \]

Once we have the quantile parts (and its inverse) trained, the weights
of the network can be optionally frozen, and the copula part is added to
the computation graph. Let the estimated quantile indices for the
\emph{ground truth} be \(\tilde{u}_i = g_{-1}(y_i,c_i(\mathbf{x}))\),
and their corresponding normal score be
\(\tilde{\mathbf{z}}^* = \Phi^{-1}(\tilde{\mathbf{u}})\), then the
Gaussian copula can be trained by minimizing the negative log
likelihood:
\[ l_3 = \mathbb{E}_{(\mathbf{x},\mathbf{y})}[2\log(|\mathbf{L}|) + (\mathbf{L}^{-1}\tilde{\mathbf{z}}^*)^\intercal (\mathbf{L}^{-1}\tilde{\mathbf{z}}^*)] + \text{const}\]

In practice, to improve learning stability by allowing better dynamic
range, and to avoid the unnecessary use of \(\Phi^{-1}(\cdot)\), we
actually replace the use of \(\mathbf{u}\) by \(\mathbf{z}^*\): instead
of quantile indices themselves, their normal scores are used as both
inputs in \(g(\cdot)\) and targets in \(g_{-1}(\cdot)\), as well as in
\(l_2\). Note \(\mathbf{u} = \Phi(\mathbf{z}^*)\) is still required to
be the weights in the QL loss function. \(\Phi(\cdot)\) has no
analytical form but is known to be well approximated by some
polynomials. \(\log(|\mathbf{L}|)\) is simply the sum of log diagonal
elements of the lower-triangle matrix, and
\(\mathbf{L}^{-1}\tilde{\mathbf{z}}^*\) can be solved by
back-substitution for the linear system
\(\mathbf{L}\tilde{\mathbf{z}} = \tilde{\mathbf{z}}^*\); all these
operations have symbolic auto-gradient implementations in common deep
learning packages, and thus the whole model can be learned using
standard gradient-based optimization. There is a numerical instability
issue: computing the inverse of \(\mathbf{L}(\mathbf{x})\), because
during training it may get initialized in an ill-conditioned state,
especially when \(d\) is large. We use the following empirical guardrail
to enforce a stable parameterization of \(\mathbf{L}\): output the
diagonal and off-diagonal elements of a raw \(\mathbf{L}\) separately;
clip the diagonal to be greater than 1; put a \(\text{tanh}(\cdot)\)
activation on off-diagonal; finally divide each row of the raw
\(\mathbf{L}\) by the row \(l^2\)-norm, so
\(\mathbf{L}\mathbf{L}^\intercal\) is a correlation matrix. This
essentially constraints the possible set of correlation matrices that
the model can learn, and works well on tested dataset. Although
implementing auto-grad regularized matrix inversion or limiting
correlation structure would be a formal solution, we describe
alternative plans of improvement in
\protect\hyperlink{discussion}{Section 5}.

\hypertarget{time-series-modeling}{%
\subsection{Time Series Modeling}\label{time-series-modeling}}

\begin{figure}
\centering
\includegraphics[width=3.33333in,height=2.5in]{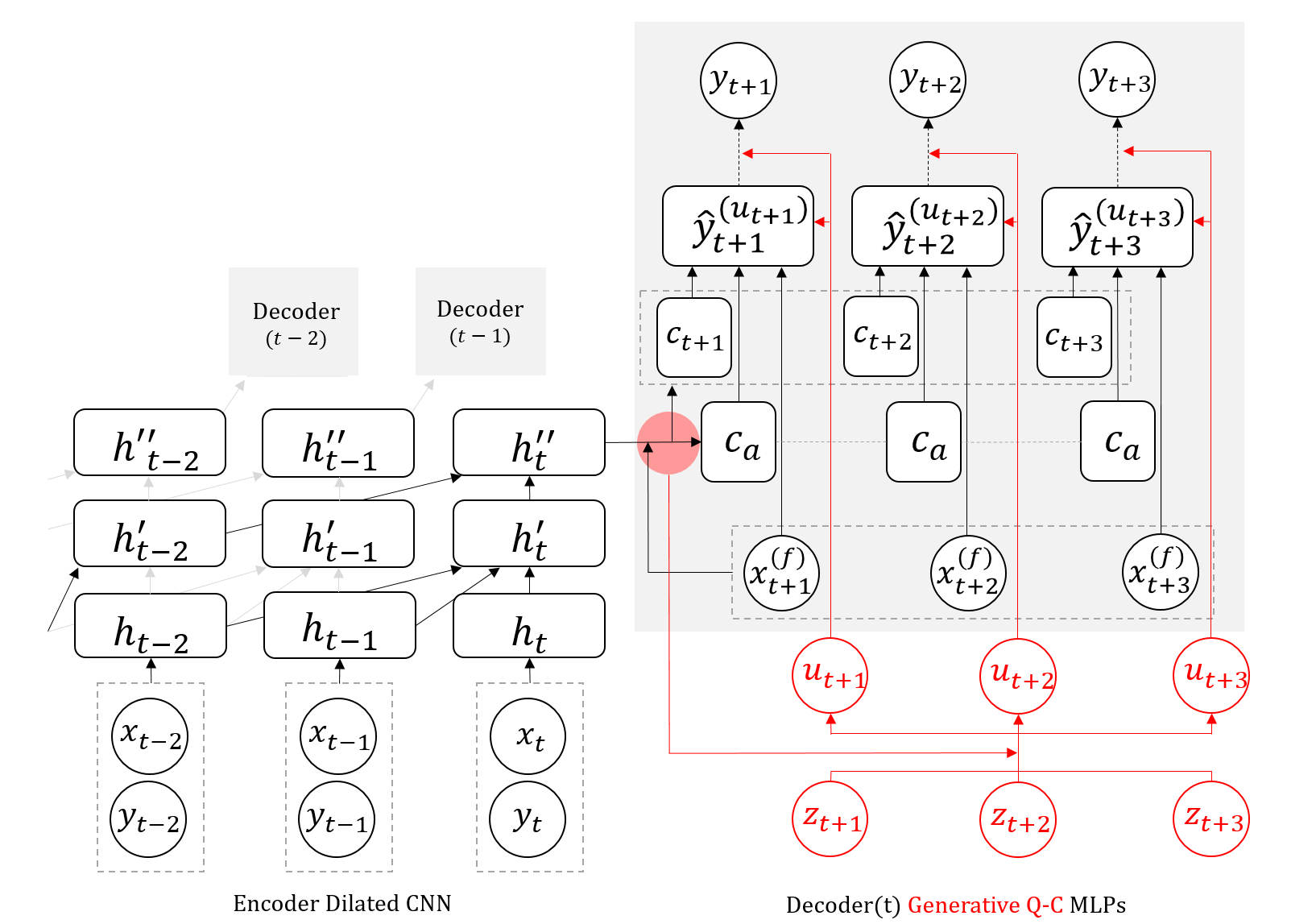}
\caption{GMQ-Forecaster: Generative Quantile-Copula model (red) applied
to MQ-CNN Forecaster (black). The red rounded shade means concatenating
all contexts \(c_{t:}\), \(c_a\) and \(x_{t:}^{(f)}\) together. Loss
functions of Copula and inverse MLPs are not shown for clarity. To
generate forecasts, either draw \(K\) random \(z_{t:}\) to get
predictive sample paths of \(y_{t:}\) and then infer any quantity of
interests using empirical statistics, or set \(u_{t:}\) to fixed numbers
in \((0,1)\) to directly fetch marginal quantile forecasts for
\(y_{t:}\).}
\end{figure}

\href{https://arxiv.org/abs/1711.11053}{Wen et al, 2017} formulated
probabilistic forecasting as a multi-target regression problem:
\(p(y_{t+1},\dots,y_{t+d}|\mathbf{x})\), where \(\mathbf{x}\) includes
past series (\(y_{:t}\)) and some other historical (\(x_{:t}\)), static
(\(x_{s}\)), and future available (\(x_{t:}^{(f)}\)) features. The
proposed MQ-forecaster framework uses a sequential net (RNN or 1D CNN)
as an encoder to process past temporal features, summarizes them into
contexts for each of the \(d\) future horizon of interests, and then
adopts multiple weight-shared MLPs as decoders to predict quantiles for
each horizon. In training, a series of decoders are \emph{forked} out of
each step in the sequential encoder to boost efficiency and stability.
The model is trained across all series, with each as a single sample.
Since it is a multi-target quantile regression, the generative
quantile-copula paradigm can be trivially added to the decoders. See
Figure 2.

\textbf{Probabilistic Forecasting and Simulation} This upgrade empowers
the MQ-forecaster with the capability to simulate a generative,
statistically consistent \emph{joint distribution} of the future time
series (any quantities of interests can be obtained by querying the
empirical statistics of a certain number of predictive sample paths),
and also to efficiently predict designated marginal quantiles for every
\(u \in (0,1)\) instead of just the pre-defined ones. The future
information \(x_{t:}^{(f)}\) (e.g.~planned promotion campaigns) can be
modified to simulate what-if action scenarios, up to some causal
inference configurations. We name the new framework Generative
Multivariate Quantile (GMQ-)forecaster.

\textbf{Cross-series association} Note
\(\tilde{\mathbf{z}} = \mathbf{L}^{-1}\tilde{\mathbf{z}}^*\) is the
implied \emph{independent} latent variables (de-correlated white noise)
of a given observation series. In the case that there are multiple time
series \(\mathbf{y}_j\), \(j = 1,\dots,M\), then the \(M\)-by-\(d\)
matrix \(\tilde{\mathbf{Z}}\) can be used to estimate the copula among
the multiple time series (either Gaussian or empirical copula). Such
estimation can also accommodate known hierarchical/similarity structure
among the series (e.g.~demand of substitutable products, inventory units
in nearby warehouses), to reduce the dimensionality. Specifically, if
the cross-series correlation matrix \(\hat{\mathbf{S}}_{M\times M}\) is
estimated from \(\tilde{\mathbf{Z}}\), subject to regularization and
sparsity constraints, then cross-series simulation can be simply
achieved by drawing latent variable matrix \(\mathbf{Z}_{M\times d}\) so
that each column follows \(N(\mathbf{0},\hat{\mathbf{S}})\), instead of
independently. Then each row of \(\mathbf{Z}\) can be fed into the model
as before to generate sample paths, and the simulation natively reflects
both cross-time and cross-series dynamics.

\textbf{Anomaly Detection} The inverse MLP outputs the implied quantile
index \(\tilde{\mathbf{u}}\), or its normal score
\(\tilde{\mathbf{z}}^*\), of the observation \(\mathbf{y}\). This comes
for free and can be conveniently interpreted as a model-based \emph{risk
score} for time series point-anomaly detection and other applications.
Likewise, multivariate anomalies (whole series) can be identified from
the density of the Gaussian Copula with \(\tilde{\mathbf{z}}^*\) and
\(\mathbf{L}\).

\hypertarget{experiment-amazon-demand-forecasting}{%
\section{Experiment: Amazon Demand
Forecasting}\label{experiment-amazon-demand-forecasting}}

We apply the GMQ-forecaster to the Amazon Demand Forecasting dataset.
180,000 products are sampled across different categories in the US
marketplace, and their weekly demand series are collected from 2014 to
2018. Available covariates include a range of suitably chosen and
standard demand drivers in three categories: history only, e.g.~past
demand units; history and future, e.g.~promotions; and static,
e.g.~product catalog fields. The 3 years of data before 2017 are used to
train the models and the rest are for evaluation. Evaluation forecasts
are created at each of the 52 weeks in 2017, while each forecast has
future horizons from 1 week to 52 weeks.

Before moving into results, we use the next sub-section to explain some
pre-requisites and conventions of both modeling and evaluating the joint
forecast distribution.

\hypertarget{mesh-gamma-and-evaluation-metrics}{%
\subsection{Mesh, Gamma and Evaluation
Metrics}\label{mesh-gamma-and-evaluation-metrics}}

\textbf{Target Interval and the Mesh} Some forecasting applications,
like demand forecasting, have a special use case: they require
distribution forecasts not only for the time series value in each future
horizon, but also for the sum of values in any future \emph{intervals}
(i.e.~consecutive horizons). Since quantile is a univariate concept and
plain Multi-Quantile nets (e.g.~MQ-CNN) only deal with marginals, the
Mesh Approach is designed as a work-around to generate distribution
forecast for any target intervals: let the maximum horizon length be
\(d\), then there are \(d\times (d-1) / 2\) possible intervals
\([t+l,t+l+s)\subset[t,t+d]\); pick a moderate subset of supporting
\((l,s)\) pairs as the \emph{mesh points}, and insert these
\(y_{[t+l,t+l+s)}\) (the total value in the interval) directly as the
multi-target of the quantile net, in addition to the horizon targets, so
each of them gets quantile forecasts; finally quantile forecasts for any
\((l,s)\) outside the mesh points are obtained by interpolating from the
nearest three mesh points (triangular interpolation). For example, a
mesh of 235 points can be used to interpolate any intervals within the
future \(d = 52\) weeks. Note that, since these mesh points are separate
targets to be optimized in the quantile net, there is no guarantee that
the probabilistic forecasts would be statistically consistent. For
example, it is possible that
\(\mathbb{E}(\hat{y}_{[t+1,t+2]}) \neq \mathbb{E}(\hat{y}_{t+1}) + \mathbb{E}(\hat{y}_{t+2})\)
or \(\hat{y}_{[t+1,t+2]}^{(u)} < \hat{y}_{t+1}^{(u)}\). This causes
difficulties if statistical inference on the joint distribution is
needed from this set of mesh point quantile forecasts.

\textbf{Gamma Fitting} Quantile nets predict a fixed set of quantiles
only. Without the knowledge of Generative Quantile Nets which learn the
whole quantile function, previous applications that require full
distribution or arbitrary quantiles usually apply interpolation or
parametric fitting on the fixed quantile predictions. For example, any
demand forecast distribution can be represented by a shifted Gamma
distribution. This is essentially a regular Gamma distribution but
shifted to the left by 1 unit, and any negative value is considered as
0. This shift is to accommodate the fact that regular Gamma has zero
probability to be exactly zero and is not practical for the
integer-value demand units of products. A quantile net could generate
P50 and P90 quantile forecasts only, and a shifted Gamma fitting
procedure estimates the two Gamma parameters from these two quantiles.
Then the parametric distribution is stored and used to predict at any
quantiles. Such procedure restricts the forecasting distribution to a
specific two-parameter family, and cannot implement the multivariate
simulation case.

\textbf{Evaluation Metrics} To evaluate the accuracy of a joint forecast
distribution for the demand forecast application, we simply follow the
same above idea and compute QL on the mesh points. Define \(QL_u(l,s)\)
as the QL of a target interval:
\(QL_u(y_{[t+l,t+l+s)},\hat{y}_{[t+l,t+l+s)}^{(u)})\), where
\(y_{[a,b)}\) is the total demand units within the time interval
\([a,b)\) and \(t+1\) is the forecast creation time (the first unknown
future point). This can be computed for each
\(l,s \in \{1, \cdots, d\}\) given \(l+s \leq d\). Although any
quantiles can be predicted, in this paper a fixed set of
\(u \in \{0.1,0.3,0.5,0.7,0.9,0.95\}\) is used for evaluation. Note that
\(QL_u(l,1)\) fully characterizes the marginal probabilistic forecast
accuracy at each horizon, while \(QL_u(l,s>1)\) is assessing some
representative slices of the joint forecast distribution. In general,
the joint distribution or simulation `realisticness' is known to be
difficult to quantify, especially for conditional models, and visual
examples of samples can be used for intuitive inspection. Apart from the
accuracy aspect, to show that MQ-forecaster with Mesh has
inconsistencies, we compute the percentage of \emph{quantile crossings}
(Q-X; a lower quantile forecast is greater than a higher one, e.g.
\(\hat{y}_{t+1}^{(0.9)} < \hat{y}_{t+1}^{(0.7)}\)) and \emph{interval
crossings} (I-X; an interval forecast is less than that of a strict
subset, e.g.
\(\hat{y}_{[t+1,t+3)}^{(0.5)} < \hat{y}_{[t+1,t+2)}^{(0.5)}\)) in
forecast instances.

\hypertarget{results}{%
\subsection{Results}\label{results}}

In this paper, we do not repeat the state-of-the-art comparisons that
already showed the superior accuracy of MQCNN, as listed in
\protect\hyperlink{introduction}{Section 1}, but instead demonstrate
that GMQ can match the accuracy of MQCNN. Candidate models include
GMQ-forecaster (\texttt{GMQ}; Quantile-Copula + MQCNN) and two settings
of MQCNN: \texttt{MQ\_mesh\_gm} predicts P50 and P90 forecasts, plus a
Gamma fitting, plus the inconsistent Mesh approach, as stated in the
previous sub-section; \texttt{MQ\_mesh\_6q} is similar but directly
predict the 6 quantiles being evaluated instead of Gamma fitting.
Another new benchmark is the latest development in the field:
Autoregressive Implicit Quantile Networks (\texttt{AIQN};
\href{https://arxiv.org/abs/1806.05575}{Ostrovski et al, 2018}). The
AIQN implementation used (Guo, 2018) is tuned for time series
forecasting and, like all other candidate models in this experiment,
uses exactly the same encoder structure as MQCNN to minimize
hyper-parameter confounding. For generative models (\texttt{GMQ} and
\texttt{AIQN}), 100 predictive samples are drawn to infer quantiles.
Finally, GMQ without copula (\texttt{GMQ\_no\_cor}) serves as a
reference assuming horizon independence, and resembles plain
MQ-forecaster without mesh.

\begin{longtable}[]{@{}llrrrrrrrr@{}}
\toprule
Targets & Model & P10QL & P30QL & P50QL & P70QL & P90QL & P95QL & Q-X &
I-X\tabularnewline
\midrule
\endhead
\((l,1)\) & \texttt{MQ\_mesh\_gm} & 1.000 & 1.000 & 1.000 & 1.000 &
1.000 & 1.000 & 0.1\% & N/A\tabularnewline
& \texttt{MQ\_mesh\_6q} & 1.000 & 1.006 & 1.006 & 1.006 & 1.018 & 1.024
& 1.8\% & N/A\tabularnewline
& \texttt{GMQ} & 1.052 & 1.017 & 1.003 & 0.994 & 1.007 & 1.023 & 0\% &
N/A\tabularnewline
& \texttt{GMQ\_no\_cor} & 1.044 & 1.006 & 1.001 & 0.999 & 1.023 & 1.085
& 0\% & N/A\tabularnewline
& \texttt{AIQN} & 1.033 & 1.110 & 1.187 & 1.301 & 1.661 & 2.002 & 0\% &
N/A\tabularnewline
\((1,s)\) & \texttt{MQ\_mesh\_gm} & 1.000 & 1.000 & 1.000 & 1.000 &
1.000 & 1.000 & 0.4\% & 8.9\%\tabularnewline
& \texttt{MQ\_mesh\_6q} & 0.942 & 0.994 & 1.005 & 1.006 & 1.024 & 1.030
& 2.4\% & 7.2\%\tabularnewline
& \texttt{GMQ} & 1.013 & 0.990 & 0.984 & 0.986 & 1.010 & 1.025 & 0\% &
0\%\tabularnewline
& \texttt{GMQ\_no\_cor} & 1.653 & 1.085 & 0.982 & 1.019 & 1.396 & 1.830
& 0\% & 0\%\tabularnewline
& \texttt{AIQN} & 1.045 & 1.108 & 1.208 & 1.377 & 1.870 & 2.302 & 0\% &
0\%\tabularnewline
\bottomrule
\end{longtable}

See Table 1 for evaluation metrics across 180K products and 52 forecast
creation times. \((l,1)\): marginal target horizons (averaging all
\(l\)); \((1,s)\): target intervals starting at forecast creation time
(averaging all \(s\)). QL values are scaled by dividing that of
\texttt{MQ\_mesh\_gm}. Q-X and I-X are for quantile crossing and
interval crossing percentages (0\% is consistent). Q-X is computed
between P50/P90 only for \texttt{MQ\_mesh\_gm}, but across all 6
quantiles for \texttt{MQ\_mesh\_6q} thus not comparable. For all
metrics, the smaller the better. The result shows that \texttt{MQ} and
\texttt{GMQ} models have comparable performance, while \texttt{AIQN}
falls short, mostly due to underbiased forecasts for longer horizons
(not shown). \texttt{GMQ\_no\_cor} has no ability to model the joint
distribution thus fails at \((1,s)\) targets at distribution tails. One
surprising fact is that the Gamma-fitted forecasts
(\texttt{MQ\_mesh\_gm}) are as accurate as the nonparametric quantile
forecasts (\texttt{MQ\_mesh\_6q}/\texttt{GMQ}) for this dataset.
\texttt{MQ} models have considerable numbers of inconsistent forecasts.
Q-crossings can be easily dealt with by sorting, but fixing I-crossings
for mesh quantiles is difficult and leave the forecast questionable when
inferences are needed, e.g.~to compute correlation between horizons.
Although metrics of only two types of aggregated target periods are
presented, the same conclusion holds for any \((l,s)\) pair.

\texttt{MQ\_mesh} models are dedicated to optimize for the mesh, which
is only one aspect of the joint distribution. \texttt{GMQ} and
\texttt{AIQN} generate sample paths to reflect the full distribution.
See Figure 3 for \texttt{GMQ} predictive simulations paths and the
corresponding quantile/correlation inference of example products. The
marginal quantiles inferred from the 100 sample paths are very close to
the direct quantiles by setting \(u_{t:}\) to the specific value (not
shown; direct quantiles can improve accuracy by 1\textasciitilde{}2\%;
also increasing from 100 to 300 samples is another \textasciitilde{}1\%
gain). The conditional copula correlation matrix depends on covariates
and is product/time-specific.

\begin{figure}
\centering
\includegraphics[width=0.5\textwidth,height=0.28\textheight]{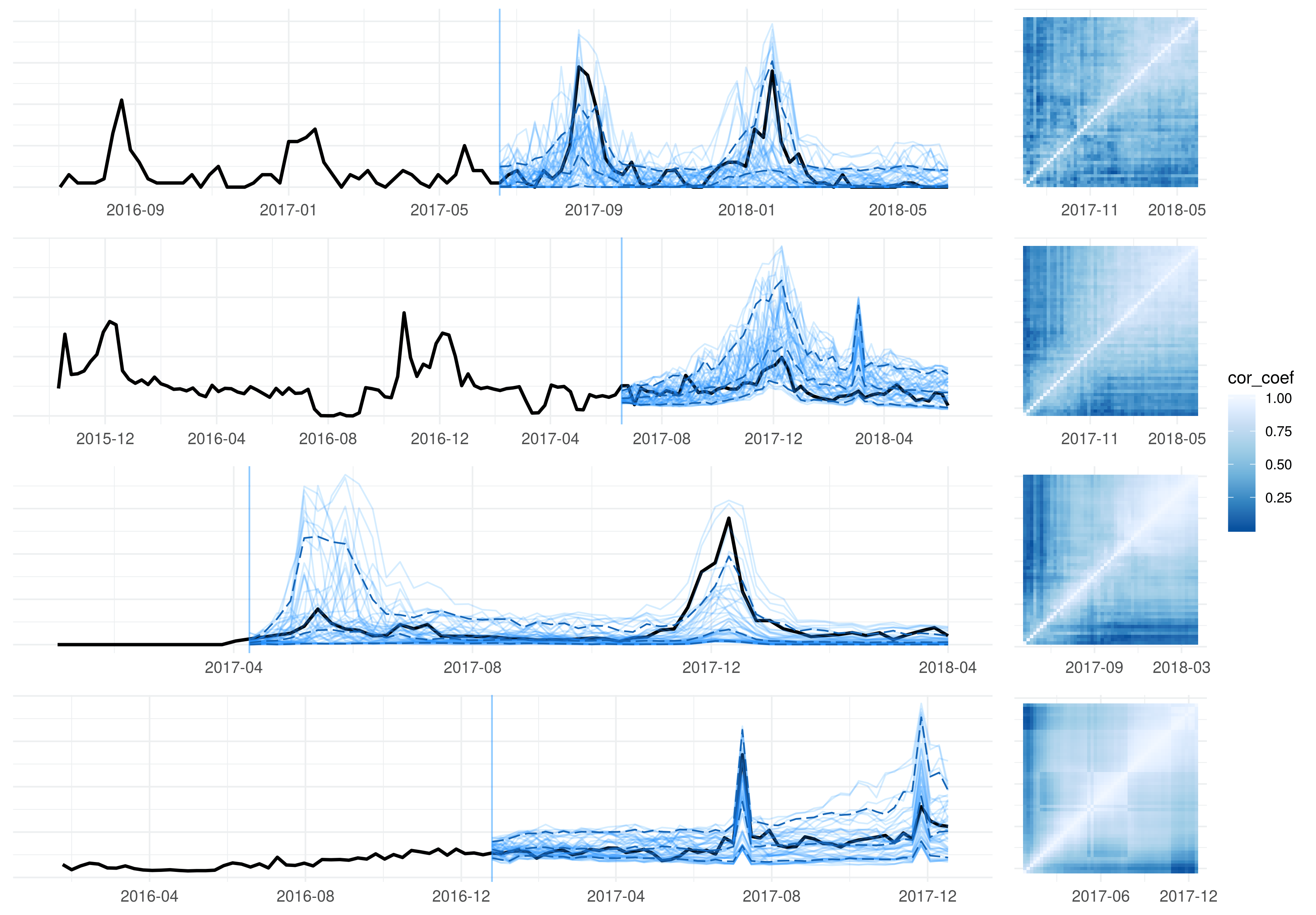}
\caption{GMQ predictive sample simulations for 4 products. Solid black
line is truth; blue vertical line is forecast creation time; solid light
blue lines are 100 predictive sample paths, each of length 52 weeks;
dashed blue lines are, from lowest to highest, marginal P10/P50/P90
inferred from samples; the horizon-by-horizon Pearson's correlation
matrix is also inferred from samples. The examples are picked to show
aspects of double/single seasonality, cold start and future event lifts,
respectively.}
\end{figure}

\hypertarget{discussion}{%
\section{Discussion}\label{discussion}}

\textbf{Related Work} Quantile-Copula decoupling has been well discussed
in statistics and forecasting (see
\href{https://www.sciencedirect.com/science/article/pii/B9780444627315000166}{Patton,
2012} for a review), but not in the space of deep or generative
modeling. \href{https://arxiv.org/abs/1610.06833}{Carlier et al, 2016}
built a connection between vector quantile regression and the optimal
transport problem, but mostly from theoretical aspects. Autoregressive
Implicit Quantile Networks (AIQN,
\href{https://arxiv.org/abs/1806.05575}{Ostrovski et al, 2018}) is
closely related to our work. AIQN pairs the univariate generative
quantile net with an autoregressive model extending to the multivariate
case, but suffers all the disadvantages of autoregressive models
(e.g.~error accumulation).
\href{https://orfe.princeton.edu/~jqfan/papers/15/TransNormal.pdf}{Fan
et al, 2016} proposed a trans-normal model by using the empirical
marginal quantiles to transform both targets and features into normal
scores, then followed by fitting a multivariate Gaussian regression. The
trans-normal model assumes simple linear relationship among the
transformed features and targets, and cannot model arbitrarily complex
interactions. Upon writing this paper, we found another similar work by
\href{https://ieeexplore.ieee.org/document/8464297}{Toubeau et al,
2019}: they used an LSTM-based forecasting model to predict quantiles,
and then a separately estimated and stored empirical copula table on a
quantile grid/cube to query scenarios. Our work differs by being a
single joint deep generative model that characterizes copulas
conditioned on different histories and covariates, while theirs assumes
an invariant copula under different contexts and depends on the choice
of a quantile grid.

\textbf{Future Work} We proposed deep generative Quantile-Copula models,
a conditional implicit generative framework that combines the marginally
expressive quantile nets and a copula generator. Minimizing marginal
quantiles loss enables various applications (e.g.~demand forecasting for
optimal inventory control), yet the model is general and can be used for
any forecasting and simulation application. The framework has much room
for extension. Both the quantile part \(g_Q(\cdot)\) and the copula part
\(g_C(\cdot)\) can be alternatively parameterized by flows-based models.
In fact, the Gaussian copula part \emph{is} a simple one-layer flow, as
the `invertible \(1\times1\) convolution' in Glow
(\href{https://arxiv.org/abs/1807.03039}{Kingma and Dhariwal, 2018}).
The invertibility of flows would yield more elegant learning and remove
the need for inverse MLPs and the Gaussian constraint on copula. This
would also help with the possible curse of dimensionality in \(d\),
where computing the inverse of \(\mathbf{L}\) becomes infeasible or not
as simple numerically. The major blocker of using flow-based models lies
in the lack of well-tested convention to condition on \(\mathbf{x}\)
while keeping the same level of model expressiveness and thus
performance - most previous work is designed for unconditional models.
Due to time limit, we leave this extension as well as the applications
outside time series data as future work.

\hypertarget{acknowledgment}{%
\subsubsection*{Acknowledgment}\label{acknowledgment}}
\addcontentsline{toc}{subsubsection}{Acknowledgment}

We would like to thank Fangjian (Richard) Guo for implementing and
experimenting the AIQN model on time series.

\hypertarget{reference}{%
\section*{Reference}\label{reference}}
\addcontentsline{toc}{section}{Reference}

Arjovsky, Martin, Soumith Chintala, and Léon Bottou. ``Wasserstein
gan.'' arXiv preprint arXiv:1701.07875 (2017).

Carlier, Guillaume, Victor Chernozhukov, and Alfred Galichon. ``Vector
quantile regression beyond correct specification.'' arXiv preprint
arXiv:1610.06833 (2016).

Dinh, Laurent, David Krueger, and Yoshua Bengio. ``NICE: Non-linear
independent components estimation.'' arXiv preprint arXiv:1410.8516
(2014).

Dinh, Laurent, Jascha Sohl-Dickstein, and Samy Bengio. ``Density
estimation using real nvp.'' arXiv preprint arXiv:1605.08803 (2016).

Fan, Jianqing, Lingzhou Xue, and Hui Zou. ``Multitask quantile
regression under the transnormal model.'' Journal of the American
Statistical Association 111, no. 516 (2016): 1726-1735.

Flunkert, Valentin, David Salinas, and Jan Gasthaus. ``DeepAR:
Probabilistic forecasting with autoregressive recurrent networks.''
arXiv preprint arXiv:1704.04110 (2017).

Gal, Yarin, and Zoubin Ghahramani. ``Dropout as a Bayesian
approximation: Insights and applications.'' In Deep Learning Workshop,
ICML, vol.~1, p.~2. 2015.

Germain, Mathieu, Karol Gregor, Iain Murray, and Hugo Larochelle.
``Made: Masked autoencoder for distribution estimation.'' In
International Conference on Machine Learning, pp.~881-889. 2015.

Gneiting, Tilmann, and Adrian E. Raftery. ``Strictly proper scoring
rules, prediction, and estimation.'' Journal of the American Statistical
Association, no. 477 (2007): 359-378.

Goodfellow, Ian, Jean Pouget-Abadie, Mehdi Mirza, Bing Xu, David
Warde-Farley, Sherjil Ozair, Aaron Courville, and Yoshua Bengio.
``Generative adversarial nets.'' In Advances in neural information
processing systems, pp.~2672-2680. 2014.

Guo, Fangjian. ``Generative modeling of time series via conditional
quantiles''. Technical report, SCOT Forecasting, Amazon, 2018.

Kingma, Diederik P., and Max Welling. ``Auto-encoding variational
bayes.'' arXiv preprint arXiv:1312.6114 (2013).

Kingma, Durk P., and Prafulla Dhariwal. ``Glow: Generative flow with
invertible 1x1 convolutions.'' In Advances in Neural Information
Processing Systems, pp.~10236-10245. 2018.

Koenker, Roger, and Gilbert Bassett Jr. ``Regression quantiles.''
Econometrica: journal of the Econometric Society (1978): 33-50.

Madeka, Dhruv, Lucas Swiniarski, Dean Foster, Leo Razoumov, Kari
Torkkola, and Ruofeng Wen. ``Sample Path Generation for Probabilistic
Demand Forecasting.'' In KDD MiLeTS workshop, 2018.

Mohamed, Shakir, and Balaji Lakshminarayanan. ``Learning in implicit
generative models.'' arXiv preprint arXiv:1610.03483 (2016).

Oord, Aaron van den, Sander Dieleman, Heiga Zen, Karen Simonyan, Oriol
Vinyals, Alex Graves, Nal Kalchbrenner, Andrew Senior, and Koray
Kavukcuoglu. ``Wavenet: A generative model for raw audio.'' arXiv
preprint arXiv:1609.03499 (2016).

Ostrovski, Georg, Will Dabney, and Rémi Munos. ``Autoregressive quantile
networks for generative modeling.'' arXiv preprint arXiv:1806.05575
(2018).

Patton, Andrew. ``Copula methods for forecasting multivariate time
series.'' In Handbook of economic forecasting, vol.~2, pp.~899-960.
Elsevier, 2013.

Sklar, Abe. ``Random variables, joint distribution functions, and
copulas.'' Kybernetika 9, no. 6 (1973): 449-460.

Toubeau, Jean-François, Jérémie Bottieau, François Vallée, and Zacharie
De Grève. ``Deep learning-based multivariate probabilistic forecasting
for short-term scheduling in power markets.'' IEEE Transactions on Power
Systems 34, no. 2 (2019): 1203-1215.

Wen, Ruofeng, Kari Torkkola, Balakrishnan Narayanaswamy. ``A
multi-horizon quantile recurrent forecaster.'' In NIPS Time Series
Workshop. (2017).

Xu, Qifa, Kai Deng, Cuixia Jiang, Fang Sun, and Xue Huang. ``Composite
quantile regression neural network with applications.'' Expert Systems
with Applications 76 (2017): 129-139.

\end{document}